\documentclass[11pt]{article}
\usepackage[margin=1in]{geometry}
\usepackage{microtype}
\usepackage[hyphens]{url}  
\usepackage{amsmath}
\usepackage{booktabs}
\usepackage{amssymb}
\usepackage{amsthm}
\newtheorem{theorem}{Theorem}

\usepackage{graphicx} 
\usepackage{natbib}  
\usepackage{caption} 
\title{Beyond Token Accuracy: Prioritizing What Matters for Visual Reconstruction}
\author{Zhicheng Liu, Zhouxiang Zhao, Chenliang Wu,\\
Zhaohui Yang\thanks{Corresponding authors.}, Zhaoyang Zhang\footnotemark[1]\\
Zhejiang University}
\date{}

\begin{document}

\maketitle

\begin{abstract}
Tokens have become a unified interface for multimodal foundation models, making visual-token communication a natural paradigm for efficient image delivery. However, existing methods typically rely on static policies that cannot jointly adapt to image content and channel conditions. Moreover, their token-level utility objectives do not necessarily translate into improved image reconstruction quality. In this paper, we propose \textbf{AdapToC}, an adaptive, reconstruction-oriented visual-token communication framework. At the transmitter, an adaptive selector jointly models image content, channel state, and communication budget to perform instance-wise resource allocation. Rather than using a fixed token rate and protection policy, it dynamically determines how many tokens should be transmitted and assigns different protection levels according to token importance and current channel conditions. At the receiver, an adaptive MaskGIT receiver incorporates channel reliability into contextual token modeling. It distinguishes tokens with different reliability levels, preserves high-confidence observations, corrects potentially corrupted tokens, and iteratively reconstructs missing content from the received evidence and global visual context. By co-designing token quantity, unequal protection, and reliability-aware recovery for image-level reconstruction quality, AdapToC achieves a peak mean PSNR gain of 4.20 dB over the strongest static baseline under matched communication costs and state-of-the-art performance among the evaluated visual-token communication methods.
\end{abstract}

\section{Introduction}
\label{sec:introduction}

Sixth-generation (6G) networks are expected to tightly integrate wireless
communication with distributed intelligence, enabling cameras, robots, and
extended-reality (XR) devices to collaborate with powerful artificial intelligence (AI) models
deployed at the network edge \cite{qiao2025tokcom,liu2026tonic}. In such end--edge systems,
resource-constrained devices must deliver visual observations over bandwidth,
latency, and energy-limited wireless links, while the available channel
quality may vary substantially over time. Transmitting raw pixels or dense
continuous features can be prohibitively expensive. Discrete visual tokens
offer a model-native alternative: a vector-quantized (VQ) encoder converts an image
into individually addressable codebook indices, which can be selectively
transmitted, unequally protected, or reconstructed by an edge generative model
\cite{oord2017vqvae,esser2021taming,hu2023robust}. This interface turns visual delivery into
a resource allocation problem, but its endpoint remains the reconstructed image
rather than the recovered token sequence.

Token accuracy is therefore an incomplete surrogate for reconstruction quality. Different token errors can have very different visual effects, depending on their locations, substituted codewords, surrounding context, and the nonlinear decoder. As a result, recovering more correct tokens does not necessarily produce a better reconstructed image.

\begin{figure}[t]
    \centering
    \includegraphics[width=\linewidth]{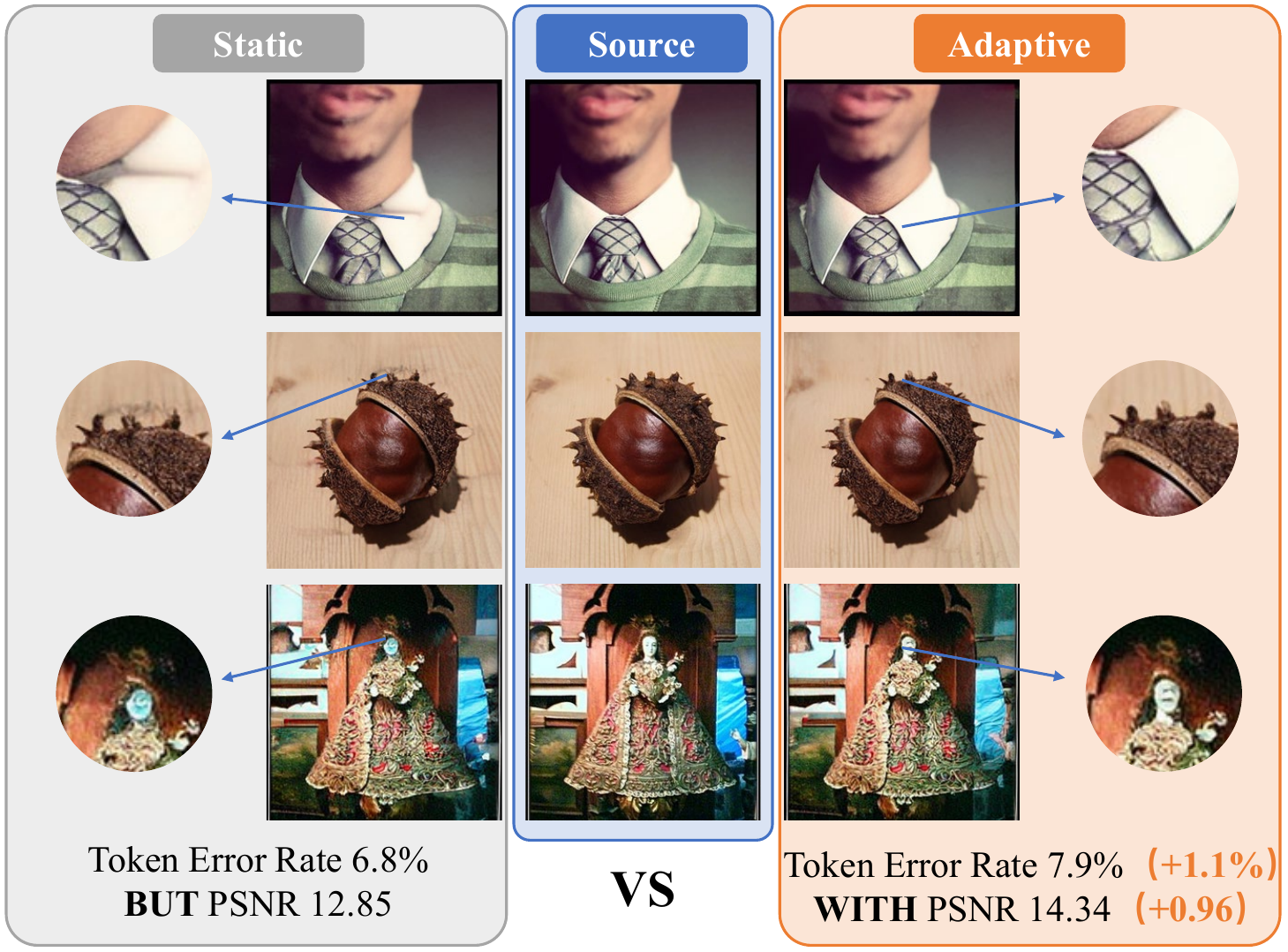}
    \caption{Examples of mismatch between token accuracy and image quality.}
    \label{fig:reversal}
\end{figure}

Fig.~\ref{fig:reversal} provides empirical witnesses of this mismatch. 
The mismatch also changes how resources should be allocated. The value of
protecting one token depends on which other tokens have already been selected,
what the receiver can infer from context, how reliably each protection tier is
delivered, and which alternative use of the remaining budget is displaced.
Moreover, a real digital link couples token actions through control coding, quadrature amplitude modulation (QAM) packing, and retransmission rules. A policy satisfying
a token-count or continuous-rate constraint may therefore be visually
suboptimal or even infeasible after the digital frame is constructed. Static
protection and independent token ranking cannot represent these interactions.

We address these problems with \textbf{AdapToC}, a reconstruction-oriented
visual-token communication framework that values complete decisions by their
terminal visual consequence. Given image content, channel state, and an exact
charged-symbol budget, its autoregressive sender constructs one complete
policy assigning DROP, WEAK, or STRONG protection across the token grid.
The contextual receiver starts from the hard-decoded token grid and its validity mask. It preserves reliable observations and adaptively intervenes on missing or unreliable positions through contextual reconstruction and refinement. This includes positions dropped by the sender or invalidated during transmission. The refined token grid is then decoded into the final image.

\begin{figure*}[t]
    \centering
    \includegraphics[width=\linewidth]{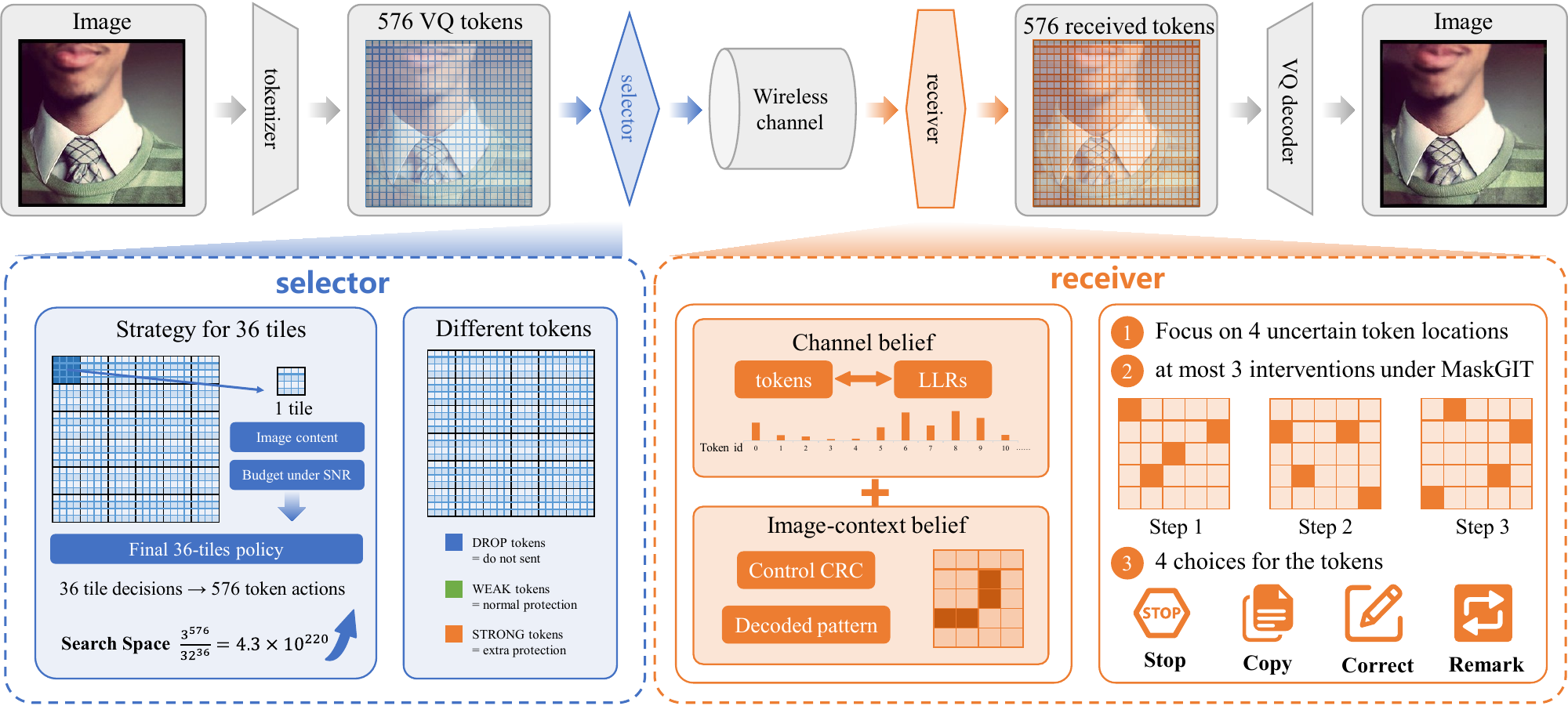}
    \caption{Overview of the end-to-end AdapToC framework.}
    \label{fig:overview}
\end{figure*}

In summary, our contributions are threefold:
\begin{itemize}
\item We develop an adaptive sender that performs complete-policy
optimization rather than independent token ranking. Conditioned on image
content, channel state, and communication budget, the sender jointly
determines token coverage and protection policy, while using an
exact digital ledger to account for control overhead, channel coding,
modulation packing, and retransmission costs.
\item We design a reliability-aware contextual receiver that combines
physical-layer evidence with global visual context. It preserves
high-confidence observations, identifies unreliable tokens, and
reconstructs missing or corrupted content through contextual completion
and refinement before image decoding.
\item We integrate the adaptive sender and contextual receiver into an
end-to-end visual-token communication system co-designed for terminal image
quality. Substantial experiments are conducted to verify the effectiveness of the proposed method, demonstrating that the
complete AdapToC system achieves the best overall performance among the evaluated visual-token communication methods.
\end{itemize}

\section{Related Work}
\label{sec:related_work}

\subsection{Discrete Visual Representations and Recovery}
VQ-VAE introduced learned discrete codebooks for representing high-dimensional
data \cite{oord2017vqvae,razavi2019vqvae2}, while VQGAN improved the perceptual quality and
compression efficiency of discrete image representations
\cite{esser2021taming,yu2024magvitv2,yu2024titok}. Discrete visual tokens were subsequently adopted for
large-scale image generation and masked image modeling by DALL-E and BEiT
\cite{ramesh2021zeroshot,bao2022beit,tian2024var,lian2026moss}. MaskGIT replaced autoregressive
generation with bidirectional masked prediction and iterative refinement
\cite{chang2022maskgit}, and Token-Critic introduced an auxiliary model for
identifying and resampling implausible generated tokens
\cite{lezama2022tokencritic}. MAGE unified representation learning and image
generation through variable-ratio masked token modeling \cite{li2023mage},
whereas Muse extended masked-token generation to text-conditioned image
synthesis \cite{chang2023muse}. More recently, Chameleon and Emu3 have treated
visual and textual tokens as a unified sequence interface for multimodal
understanding and generation \cite{team2024chameleon,wang2024emu3}. Collectively,
these studies establish discrete tokenization and contextual prediction as
effective tools for visual representation and recovery, although they generally
assume clean token inputs rather than observations affected by communication
errors and heterogeneous reliability.

\subsection{Image Transmission Over Noisy Channels}
Deep joint source--channel coding (JSCC) maps images directly to continuous channel
symbols and jointly optimizes transmission and reconstruction over a
differentiable channel \cite{bourtsoulatze2019deepjscc}. Subsequent work has
extended this framework to feedback-assisted transmission
\cite{kurka2020deepjsccf}, bandwidth-agile coding
\cite{kurka2021bandwidth}, and signal-to-noise ratio (SNR)-adaptive feature modulation
\cite{xu2022adjscc}. Channel-adaptive JSCC has also been integrated with orthogonal frequency division multiplexing (OFDM)
to address multipath fading and practical physical-layer structures
\cite{wu2022channeladaptive,yang2022ofdmguided}. NTSCC combines nonlinear transform coding,
entropy modeling, and JSCC to support content-dependent rate allocation
\cite{dai2022ntscc}. Transformer-based architectures, including WITT and
SwinJSCC, improve long-range visual modeling and support adaptation across
channel conditions and transmission rates
\cite{yang2023witt,yang2024swinjscc}. DeepJSCC-l++ further enables a single
model to operate across multiple bandwidth ratios and SNRs
\cite{bian2023deepjscclpp}. In the digital domain, JCM-VAE learns discrete
coding and modulation jointly through a variational formulation
\cite{bo2024jcm}. These approaches primarily optimize global image latents,
feature maps, or channel-symbol sequences, with communication resources usually
controlled at the representation or channel level rather than at the level of
individual visual tokens \cite{zhang2025hjscc,erdemir2023generativejscc,wu2023deepjsccmimo,wu2024cddm,wang2025diffcom,zhang2025mdjcm,zhang2026sgdjscc}.

\section{Method}
\subsection{Problem Formulation}
\label{sec:problem}

A frozen VQ encoder maps an input image $x$ into a $24\times24$ grid of
discrete visual tokens,
\begin{equation}
z=E(x)=(z_1,\ldots,z_N),
\qquad
N=24\times24=576.
\end{equation}
The tokenizer uses a codebook containing
$K=16{,}384$ learned visual codewords. Each token
$z_i\in{0,\ldots,K-1}$ is an integer index identifying one entry in this
shared codebook, rather than a continuous feature vector.
After transmission, the receiver produces a recovered token grid
$\hat{z}=(\hat{z}_1,\ldots,\hat{z}_N)$. The corresponding codebook entries
are then mapped back to the image domain by the frozen VQ decoder, $\hat{x}=D(\hat{z})$.
The encoder, codebook, and decoder are shared by the sender and receiver and
remain frozen throughout the training of AdapToC.

We divide the $24\times24$ token grid into $M=36$ non-overlapping
$4\times4$ tiles. A direct token-wise decision would assign
DROP, WEAK, or STRONG independently to
each of the 16 tokens in a tile, resulting in $3^{16}$ possible assignments.
This action space is unnecessarily large and difficult to reconcile with the
discrete communication budget. We therefore use a shared and frozen table of
32 tile-level transmission patterns, denoted by K32. Each K32 entry specifies
which tokens are omitted and which are transmitted through the WEAK or STRONG
protection stream. The sender selects one K32 entry for each tile, and the
resulting 36 tile decisions form a complete transmission policy
\begin{equation}
a=(a_1,\ldots,a_M), \qquad a_j\in\{1,\ldots,32\}.
\end{equation}
K32 thus provides a compact and protocol-compatible local action space, while
the exact table entries are fixed across training and evaluation and reported
in the supplementary material. Sender-side image features used to select
$a$ are not transmitted.

\subsubsection{Exact Communication Budget}
The executable cost oracle follows the actual digital protocol. It includes the
216-bit control message, forward error correction (FEC) expansion, integer modulation packing, and the
separate packing states of the WEAK and STRONG streams. The charged-symbol cost
of a complete policy is
\begin{equation}
C(a)=C_{\mathrm{ctrl}}+C_{\mathrm{weak}}(a)+C_{\mathrm{strong}}(a).
\label{eq:policy-cost}
\end{equation}
We use only legal policies that consume all the available budget exactly:
\begin{equation}
\mathcal{A}_{B}
=
\{a:\mathrm{Legal}(a)=1,\ C(a)=B\}.
\label{eq:feasible-set}
\end{equation}
No padding symbols are inserted. Since $C_{\mathrm{weak}}$ and
$C_{\mathrm{strong}}$ are computed after the actual FEC and modulation packing
operations, $C(a)=B$ describes the final physical-link occupancy rather than a
token-count approximation.

\subsubsection{Terminal Reconstruction Objective}
For SNR $\gamma$ and channel realization $\omega$, let
$H_{\gamma}(z,a,\omega)$ denote the complete digital link, including
modulation, FEC, channel corruption, decoding, cyclic redundancy check (CRC) status, and soft reliability
evidence. A receiver $R$ maps this evidence to a recovered token grid. We
evaluate a complete policy only after receiver inference and image decoding:
\begin{equation}
V_x(a;\gamma,R)
=
\mathbb{E}_{\omega}
\left[
U\left(x,D\left(R\left(H_{\gamma}(z,a,\omega)\right)\right)\right)
\right],
\label{eq:terminal-value}
\end{equation}
where $U(x,\hat{x})$ denotes the utility function. The sender-training pipeline evaluates policies with a frozen direct receiver
$R_0$ and approximates
\begin{equation}
a_0^{*}(s)
=
\arg\max_{a\in\mathcal{A}_{B}}V_x(a;\gamma,R_0),
\qquad s=(x,\gamma,B).
\label{eq:sender-objective}
\end{equation}

\subsection{Proposed AdapToC}
\label{sec:method}

AdapToC contains an adaptive sender and a contextual receiver, as depicted in Fig.~\ref{fig:overview}. The sender
allocates the communication budget across image regions and protection levels.
The receiver uses channel reliability and visual context to recover missing or
uncertain tokens.

\subsubsection{Why Complete Policies?}
\label{sec:design-theory}

Token accuracy counts correct indices but does not measure their effect on the
reconstructed image. That effect depends on spatial position, codeword
substitution, surrounding tokens, and the nonlinear decoder. A grid with more
correct tokens can therefore still produce a worse image.

Fix a source grid $z$ and decoder-level loss $L_z(\hat{z})$. The set of
predictions with exactly $r$ token errors is
\begin{equation}
\mathcal{S}_{r}(z)
=
\{\hat{z}:d_H(z,\hat{z})=r\}.
\label{eq:hamming-shell-set}
\end{equation}
The minimum and maximum losses within this Hamming shell are
\begin{equation}
\underline{L}_{r}
=
\inf_{\hat{z}\in\mathcal{S}_{r}(z)}L_z(\hat{z}),
\qquad
\overline{L}_{r}
=
\sup_{\hat{z}\in\mathcal{S}_{r}(z)}L_z(\hat{z}).
\label{eq:hamming-shell-range}
\end{equation}

\begin{theorem}
    Token accuracy does not generally order reconstruction quality.
\end{theorem}

Token accuracy orders decoder-level loss for every pair of predictions if and
only if $\overline{L}_{r}\leq\underline{L}_{r+1}$ for every pair of adjacent
nonempty Hamming shells. If two adjacent shells overlap, a prediction with more
correct tokens can have larger reconstruction loss.

The proof is provided in the supplement. At the transmission level, the value
of one decision also depends on which other regions are sent, while its cost
depends on the current packing state. AdapToC therefore learns from terminal
outcomes of complete, executable policies rather than independent token labels
or additive token scores.

\subsubsection{Ledger-Aware Adaptive Sender}
\label{sec:actor}

A learned priority head first orders the 36 tiles as $\tau$. The actor then
selects one K32 pattern at a time:
\begin{equation}
\pi_{\theta}(a\mid s)
=
\prod_{t=1}^{M}
\pi_{\theta}
\left(a_{\tau_t}\mid s,a_{\tau_{<t}},\ell_t\right).
\label{eq:actor-factorization}
\end{equation}
The protocol ledger is
\begin{equation}
\ell_t
=
\left(
C_t^{\mathrm{rem}},
 r_t^{\mathrm{W}},
 r_t^{\mathrm{S}},
 n_t^{\mathrm{W}},
 n_t^{\mathrm{S}},
 h_t^{\mathrm{sel}}
\right).
\label{eq:ledger-state}
\end{equation}
It records the remaining symbols, WEAK and STRONG packing remainders, payload
sizes, and a summary of previous selections. Each decision therefore conditions
on image content, continuous SNR, the total budget, previously selected
patterns, and the current digital packing state.

Before selecting a pattern, a viability mask retains only choices that admit at
least one legal exact-budget completion:
\begin{equation}
\mathcal{V}_{t}(s,a_{\tau_{<t}})
=
\left\{\nu:
\begin{array}{l}
\exists a_{\tau_{>t}}\ \mathrm{such\ that}\\
(a_{\tau_{<t}},\nu,a_{\tau_{>t}})\in\mathcal{A}_{B}
\end{array}
\right\}.
\label{eq:viability-kernel}
\end{equation}

If $a_{\tau_t}\in\mathcal{V}_{t}(s,a_{\tau_{<t}})$ at every step, the completed
policy belongs to $\mathcal{A}_{B}$ and therefore satisfies $C(a)=B$ without
padding.
The charged-symbol budget is thus enforced while the policy is generated,
including integer packing effects that a token-count constraint would miss.

\subsubsection{Complete-Policy Supervision}
\label{sec:training}

Greedy actor decoding first produces an anchor policy $a^{(0)}$. A deterministic
C13 operator then constructs complete alternatives using a fixed
$1/2/2/4/2/2$ allocation: one anchor, two release variants, two investment
variants, four exchange variants, two variants that expand spatial coverage,
and two variants that concentrate protection. These operations redistribute the
fixed budget across image regions and protection levels. After each edit, the
remaining tile decisions are completed with legal K32 patterns so that the
resulting policy still satisfies $C(a)=B$. A proposal is discarded when no
exact-budget completion exists. Every completed proposal then passes through
the cost oracle, viability check, and deduplication. The resulting candidate set
satisfies
\begin{equation}
\mathcal{C}(s)\subseteq\mathcal{A}_{B},
\qquad
|\mathcal{C}(s)|\leq13.
\label{eq:c13-set}
\end{equation}
The exact operators and frozen K32 manifest are listed in the supplement.

All candidates are evaluated using the same $K$ channel realizations. Paired
draws reduce comparison noise because every policy experiences identical
channel conditions. With the frozen direct receiver $R_0$, the reconstruction
for candidate $a$ under draw $k$ is
\begin{equation}
\hat{x}_{k}(a)
=
D\left(R_0\left(H_{\gamma}(z,a,\omega_k)\right)\right).
\label{eq:paired-reconstruction}
\end{equation}
Its metric vector is
\begin{equation}
\mathbf{q}_{k}(a)
=
\mathbf{q}\left(x,\hat{x}_{k}(a)\right),
\label{eq:paired-metrics}
\end{equation}
where the entries are PSNR, SSIM, negative LPIPS, and CLIP. Let $u_k(a)$ be the
fixed scalar training utility computed from this vector. Relative to the anchor,
a candidate is accepted only when
\begin{equation}
\Delta q_{k,m}(a)\geq \epsilon_m
\qquad
\forall k,m,
\label{eq:metric-safety}
\end{equation}
and
\begin{equation}
\Delta u_k(a)>\eta
\qquad
\forall k,
\label{eq:utility-gain}
\end{equation}
where $m$ denotes the metric index, $\epsilon_m$ and $\eta$ are positive threshold, respectively. Let $\mathcal{C}_{+}(s)$ contain the candidates that pass both conditions, and
define their mean utility as
\begin{equation}
\bar{u}(a)=\frac{1}{K}\sum_{k=1}^{K}u_k(a).
\label{eq:mean-utility}
\end{equation}
When $\mathcal{C}_{+}(s)$ is nonempty, the teacher selects
\begin{equation}
a_{\mathrm{tea}}
=
\arg\max_{a\in\mathcal{C}_{+}(s)}\bar{u}(a).
\label{eq:teacher-selection}
\end{equation}
Otherwise, it retains the anchor $a^{(0)}$. The metric-wise constraints prevent
large regressions, while the scalar utility only ranks candidates that have
passed those constraints.

The actor distills $a_{\mathrm{tea}}$ using autoregressive negative
log-likelihood. The same records train an auxiliary value model to predict
terminal metrics over the bounded candidate support. C13, the teacher, and the
value model are used only during training. Values of $K$, $\epsilon_m$, $\eta$,
and the fixed utility scaling are reported in the implementation details.

\begin{figure*}[t]
    \centering
    \includegraphics[width=\linewidth]
    {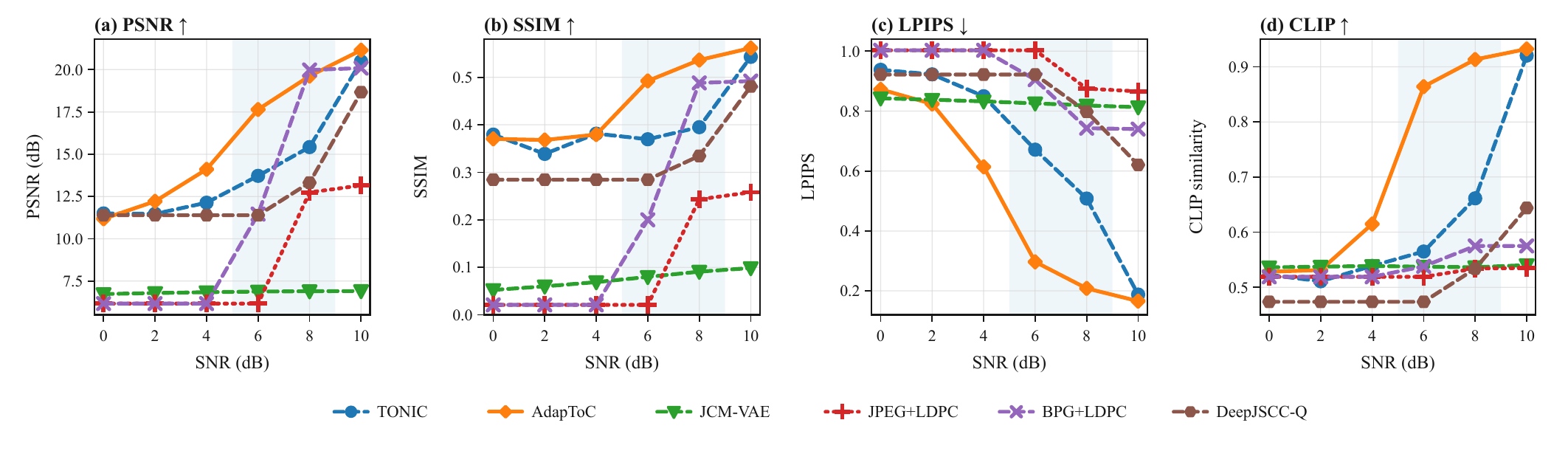}
    \caption{Comparison with digital communication baselines across
    channel conditions at $B=6052$.}
    \label{fig:all-baselines}
\end{figure*}

\subsubsection{Reliability-Aware Contextual Receiver}
\label{sec:receiver}

For token position \(i\), let \(\mathcal{E}_{\gamma,i}\) denote the
physical-layer evidence available at the receiver under SNR \(\gamma\). It
contains the decoded token index, when available; soft-demodulation and FEC
information such as log likelihood ratios (LLRs); CRC or protocol-validity status; and a reliability
descriptor \(\rho_i\). The descriptor \(\rho_i\) summarizes how strongly the
physical observation at position \(i\) should be trusted.

The physical-layer evidence induces a channel distribution over the \(K\)
codebook entries:
\begin{equation}
p_{\mathrm{ch}}^{(i)}(c)
\triangleq
\Pr\!\left(
    z_i = c
    \mid
    \mathcal{E}_{\gamma,i}
\right),
\qquad
c \in \{1,\ldots,K\},
\label{eq:channel-evidence}
\end{equation}
where \(z_i\) is the unknown ground-truth token at position \(i\), and
\(p_{\mathrm{ch}}^{(i)}\) represents the receiver's token belief derived
solely from the physical channel.

The receiver maintains an intermediate token grid
\[
\hat{\mathbf z}^{(t)}
=
\bigl(
    \hat z_1^{(t)},\ldots,\hat z_N^{(t)}
\bigr)
\]
at iteration \(t\). Let
\(\mathcal M_t \subseteq \{1,\ldots,N\}\) denote the positions that remain
masked or are selected for reconsideration, and let
\[
\overline{\mathcal M}_t
=
\{1,\ldots,N\}\setminus\mathcal M_t
\]
denote the currently visible positions. The corresponding visible tokens are
written as
\[
\hat{\mathbf z}_{\overline{\mathcal M}_t}^{(t)}
\triangleq
\left\{
    \hat z_j^{(t)}
    : j \in \overline{\mathcal M}_t
\right\}.
\]

Conditioned on these visible tokens, the contextual model predicts
\begin{equation}
p_{\mathrm{ctx}}^{(t,i)}(c)
\triangleq
\Pr_{\psi}\!\left(
    z_i = c
    \mid
    \hat{\mathbf z}_{\overline{\mathcal M}_t}^{(t)}
\right),
\qquad
c \in \{1,\ldots,K\},
\label{eq:contextual-distribution}
\end{equation}
where \(p_{\mathrm{ctx}}^{(t,i)}\) describes which codebook entries are
visually plausible at position \(i\) under the current reconstruction
context.

The channel and contextual distributions are then combined according to the
physical reliability:
\begin{equation}
p_{\psi}^{(t,i)}(\cdot)
\triangleq
\operatorname{Fuse}_{\psi}\!\left(
    p_{\mathrm{ch}}^{(i)}(\cdot),
    p_{\mathrm{ctx}}^{(t,i)}(\cdot),
    \rho_i
\right).
\label{eq:posterior-fusion}
\end{equation}

The resulting \(p_{\psi}^{(t,i)}\) is the fused posterior distribution over
the \(K\) codebook entries. A reliable physical observation places greater
weight on \(p_{\mathrm{ch}}^{(i)}\), whereas a missing or uncertain
observation relies more strongly on \(p_{\mathrm{ctx}}^{(t,i)}\). The
receiver uses this posterior to estimate token confidence and choose among
COPY, CORRECT, and REMASK.
Reliability evidence determines how strongly the receiver trusts the channel
decision relative to visual context. Protocol validity alone does not make a
hard decision unquestionably correct.

Recovery has two stages. Three scheduled contextual-completion rounds first
fill positions that were dropped or unusable after channel decoding. A bounded
controller may then perform at most $T_{\max}=3$ interventions:
\begin{equation}
o^{(t)}
\in
\{\mathrm{STOP},\mathrm{COPY},\mathrm{CORRECT},\mathrm{REMASK}\}.
\label{eq:receiver-actions}
\end{equation}
STOP ends refinement. COPY preserves reliable observations. CORRECT replaces a
suspicious hard decision using the fused posterior. REMASK removes a
low-confidence token so that it can be estimated again from context. The final
token grid is decoded once by $D$. During training, the clean grid is used only
as the target. At inference, $R_{\mathrm{ctx}}$ uses only channel- and
protocol-observable evidence.

\subsubsection{Training and Deployment}
\label{sec:deployment}

Sender targets are
generated by evaluating C13 candidates with the frozen direct receiver $R_0$.
The contextual receiver $R_{\mathrm{ctx}}$ is trained to recover tokens from
missing, corrupted, and reliability-annotated channel observations.

At deployment, the adaptive sender performs one actor pass, selects one viable
K32 pattern for each tile, and executes one exact-budget digital transmission.
The contextual receiver then runs three completion rounds and at most three
optional interventions before the recovered token grid is decoded once by the
shared VQ decoder. Candidate generation, paired channel trials, teacher
selection, and the auxiliary value model are used only during training.

\section{Experiments}
\label{sec:experiments}


\begin{figure*}[t]
    \centering
    \includegraphics[width=0.9\linewidth]
    {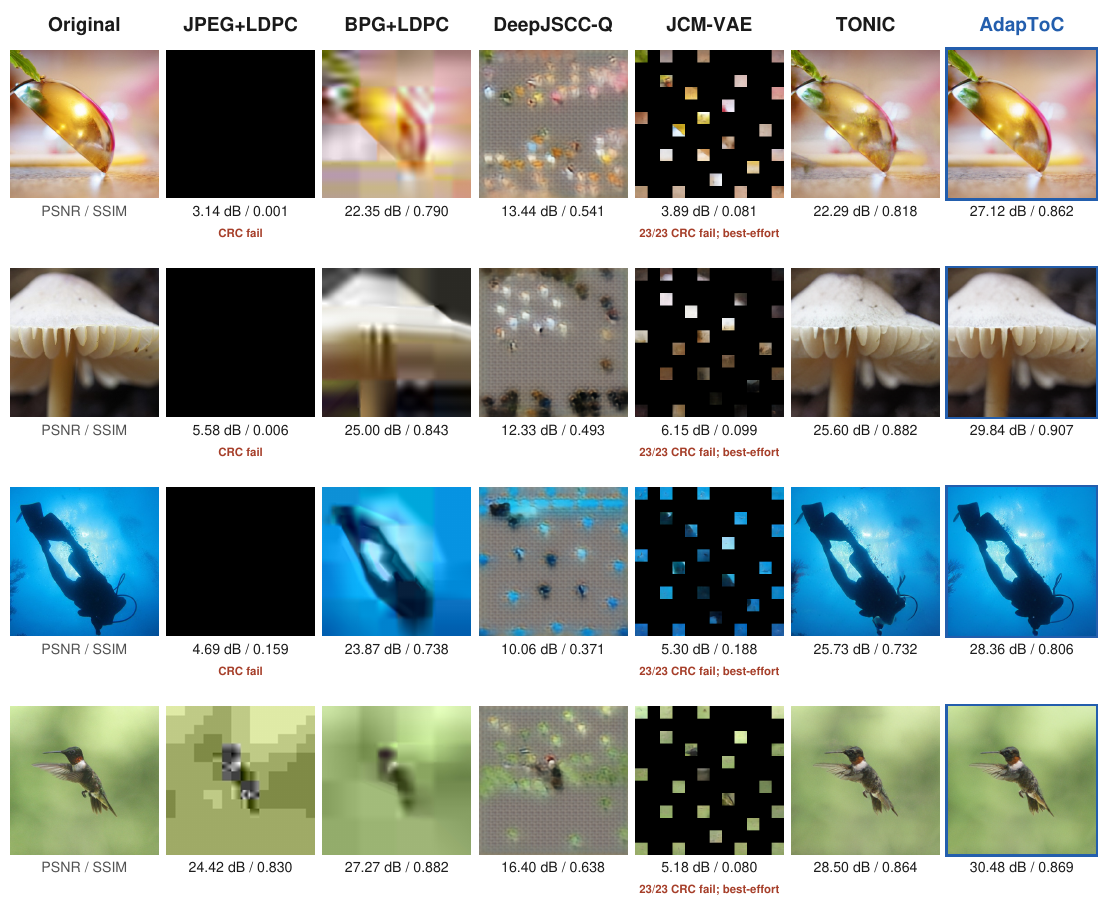}
    \caption{Qualitative comparison at SNR $=8$ dB and $B=6052$.}
    \label{fig:qualitative-8db}
\end{figure*}

\subsection{Experimental Setup}
\label{sec:exp-setup}

\subsubsection{Dataset and Representation}
We evaluate 512 held-out ImageNet-100 images \cite{deng2009imagenet} at $384\times384$ resolution. A
frozen VQGAN tokenizer \cite{esser2021taming} maps each image to a $24\times24$ grid of 576 discrete
indices from a $2^{14}$-entry codebook. One unified AdapToC checkpoint covers
$\gamma\in\{0,2,4,6,8,10\}$ dB and exact charged-symbol budgets
$B\in\{5632,6052\}$. This is enabled by the proposed image--channel adaptive
policy: the sender conditions on the source image, current SNR, and budget to
select the transmitted tokens and their protection levels at run time, rather
than requiring a separately trained model for each operating point.

\subsubsection{Controlled Evaluation and Metrics}
The controlled $2\times2$ evaluation crosses ``TONIC'' and AdapToC senders with
the direct receiver $R_0$ and contextual completion $R_{\rm ctx}$. All four
arms share image order, paired channel draws, tokenizer, decoder, LDPC/QAM
implementation, control format, and exact hardcap accounting. The receiver
$R_0$ is not an unconditional hard-token receiver: under a fixed utility
configuration, it retains a PHY hard decision only when accepting that token
is preferred to erasing it; otherwise the position is marked missing.
Therefore, ``direct'' means that $R_0$ performs no generative completion, not
that it accepts every hard decision. The contextual receiver starts from the
same utility-gated grid, preserves the accepted physical evidence, and fills
the erased positions using visual context before the shared VQ decode.

We report PSNR, SSIM, LPIPS, and CLIP similarity; higher is better except for
LPIPS \cite{wang2004ssim,zhang2018lpips,radford2021clip}. Unless stated otherwise, aggregate values are equal-cell means over the
12 SNR--budget conditions.

\subsubsection{Baselines}
All evaluated baselines use executable digital transmission pipelines. ``TONIC''
represents fixed visual-token allocation and protection \cite{liu2026tonic}; ``JCM-VAE'' and
``DeepJSCC-Q'' provide learned digital references \cite{bo2024jcm,tung2022deepjsccq}; and ``JPEG+LDPC'' and
``BPG+LDPC'' represent conventional separation-based digital systems. Together
they cover fixed token transmission, learned digital coding, and traditional
source/channel separation.

\begin{figure*}[t]
    \centering
    \includegraphics[width=\linewidth]
    {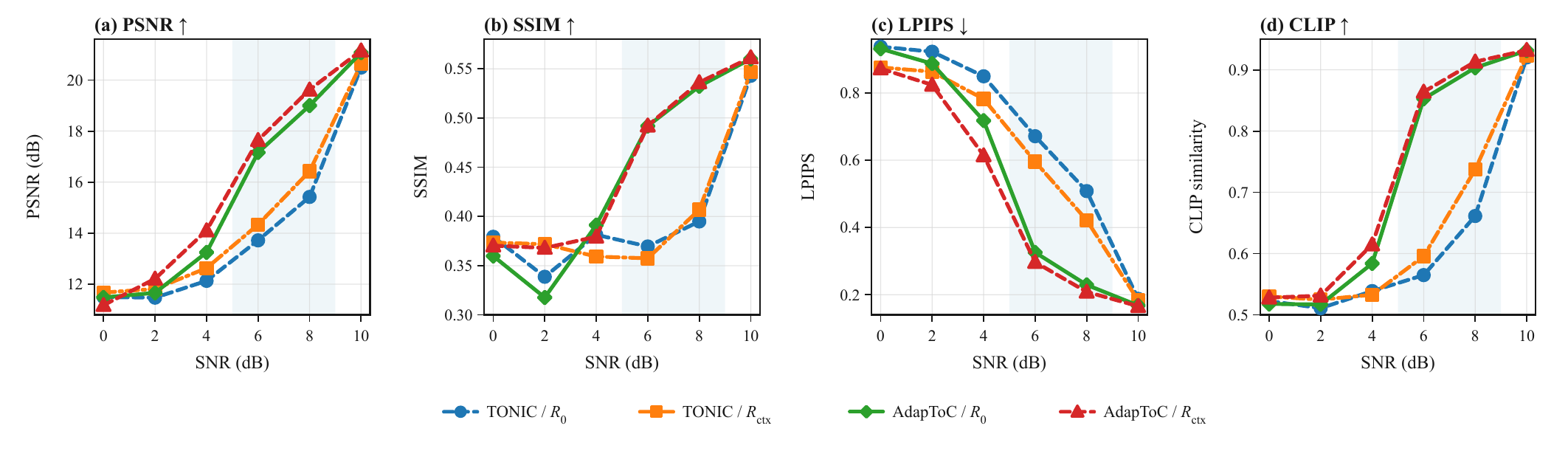}
    \caption{Sender--receiver ablation ablation at $B=6052$.}
    \label{fig:sender-receiver-ablation}
\end{figure*}

\subsection{Comparisons and Performance Analysis}
\label{sec:performance-analysis}

\subsubsection{Quantitative Analysis}

Fig.~\ref{fig:all-baselines} establishes the overall performance landscape.
Averaged equally over the six SNR conditions, AdapToC achieves 15.986 dB PSNR,
0.4513 SSIM, 0.4970 LPIPS, and 0.7307 CLIP, compared with 14.122 dB, 0.4011,
0.6796, and 0.6198 for ``TONIC''. AdapToC therefore improves all four aggregate
metrics. ``TONIC'' remains the strongest baseline but cannot reallocate its fixed
token and protection pattern to the current image and channel. ``JCM-VAE'' changes
smoothly with SNR but remains at a low reconstruction level, while ``DeepJSCC-Q''
provides gradual digital degradation without matching AdapToC. ``JPEG+LDPC'' and
``BPG+LDPC'' exhibit the expected waterfall: quality rises sharply once channel
decoding succeeds, but low-SNR frame failures reduce their across-condition
performance.


\subsubsection{Qualitative Comparisons}

Fig.~\ref{fig:qualitative-8db} visualizes four representative cases at the
operating point where the quantitative advantage is close to its maximum.

In the four displayed cases, AdapToC improves PSNR over ``TONIC'' by 4.83, 4.24,
2.63, and 1.98 dB, respectively. ``JPEG+LDPC'' suffers complete or severe frame-level
failures in several examples; ``BPG+LDPC'' preserves coarse content but loses fine
structure; ``DeepJSCC-Q'' introduces visible texture and structural artifacts; and
``JCM-VAE'' produces low-fidelity best-effort outputs. ``TONIC'' reconstructs the main
semantic objects, whereas AdapToC more accurately preserves their boundaries,
shape, and low-frequency appearance.

\subsubsection{Beyond Token Accuracy}
Token accuracy counts exact codebook-index matches but does not measure their
visual consequences. In seven registered cases, AdapToC
has 0.17--3.30 percentage points lower token accuracy than ``TONIC'', yet
improves PSNR by 0.189--0.962 dB, SSIM by 0.0031--0.0263, and CLIP by
0.0033--0.0485, while reducing LPIPS by 0.0010--0.0212. These reversals show
that recovering more exact token identities does not necessarily produce the
better terminal image, supporting reconstruction-oriented allocation and
recovery rather than token accuracy as the sole objective.

\subsection{Ablation Study}
\label{sec:module-ablation}

Under the shared $R_0$, replacing the ``TONIC'' sender with AdapToC improves the
mean by 1.4218 dB PSNR, 0.03762 SSIM, 0.14008 LPIPS reduction,
and 0.09832 CLIP, as shown in Fig.~\ref{fig:sender-receiver-ablation}. Because both arms use the same utility-gated acceptance and
erasure rule, this contrast isolates the sender policy. With the AdapToC sender
fixed, $R_{\rm ctx}$ adds 0.4310 dB PSNR, 0.01194 SSIM, 0.05153 LPIPS reduction,
and 0.01586 CLIP. The sender accounts for 76.7\% of the complete 1.8528 dB
PSNR gain and is the dominant component. The additional PSNR gain from the
contextual receiver is smaller, but its LPIPS and CLIP improvements show a
complementary perceptual benefit from completing positions erased by $R_0$.

\subsection{Hardware-Aware Evaluation}
\label{sec:hardware-evaluation}

AdapToC evaluates alternative complete policies only during training. Online
execution uses one sender actor to select one digital transmission policy,
followed by contextual completion and one terminal VQ decode.

\begin{table}[t]
    \centering
    \begingroup
    \footnotesize
    \setlength{\tabcolsep}{5.0pt}
    \renewcommand{\arraystretch}{1.08}
    \begin{tabular}{@{}lrr@{}}
        \toprule
        Online component & GFLOPs & Share \\
        \midrule
        Sender actor & 0.7030 & 0.116\% \\
        Selected digital PHY & 0.0158 & 0.003\% \\
        Contextual completion & 34.390 & 5.679\% \\
        Terminal VQ decode & 570.465 & 94.202\% \\
        \midrule
        \textbf{Total} & \textbf{605.574} & \textbf{100.000\%} \\
        \bottomrule
    \end{tabular}
    \endgroup
    \caption{Online computation breakdown.}
    \label{tab:online-compute}
\end{table}

\begin{table}[t]
    \centering
    \begingroup
    \footnotesize
    \setlength{\tabcolsep}{2.5pt}
    \renewcommand{\arraystretch}{1.08}
    \begin{tabular}{@{}lrrrrr@{}}
        \toprule
        Metric & Ref. & $\delta^{\star}$ & Worst & $\Delta$ & Rel. $\Delta$ \\
        \midrule
        PSNR $\uparrow$  & 15.646 & $+2$ & 15.385 & $-0.2610$ & $-1.67\%$ \\
        SSIM $\uparrow$  & 0.4130 & $+2$ & 0.4101 & $-0.0029$ & $-0.70\%$ \\
        LPIPS $\downarrow$ & 0.5346 & $-2$ & 0.5424 & $+0.0078$ & $+1.46\%$ \\
        CLIP $\uparrow$  & 0.7597 & $-2$ & 0.7539 & $-0.0058$ & $-0.76\%$ \\
        \bottomrule
    \end{tabular}
    \endgroup
    \caption{Robustness to sender-side SNR-estimation error.}
    \label{tab:snr-mismatch}
\end{table}

As illustrated in Table~\ref{tab:online-compute}, the sender actor contributes only 0.116\% of the counted online computation;
the shared frozen VQ decoder remains the dominant component.
Table~\ref{tab:snr-mismatch} depicts the performance of AdapToC under SNR mismatch.
For each metric, $\delta^{\star}$ denotes the most adverse offset in $\{-2,-1,-0.5,0,0.5,1,2\}$ dB.
The largest adverse PSNR change is 0.2610 dB, substantially smaller than the 1.8528 dB matched
average gain. Optimistic SNR estimates are more harmful to pixel fidelity,
whereas conservative estimates mainly affect perceptual metrics.

\section{Conclusion}
This paper presents AdapToC, an adaptive visual-token communication framework that shifts the optimization objective from proxy token accuracy to terminal reconstruction fidelity. By co-designing a ledger-aware sender for complete-policy resource allocation and a contextual receiver that fuses physical-layer reliability with visual context, AdapToC dynamically prioritizes visually critical tokens. Ultimately, this work highlights the importance of task-oriented resource allocation rather than uniform proxy metrics for future multimodal systems.

\raggedbottom
\bibliographystyle{plainnat}
\bibliography{references}


\end{document}